\documentclass{llncs}

\usepackage{color}

\usepackage{amsmath, amssymb}
\usepackage{graphicx}

\usepackage{xspace}
\usepackage{ifthen}
\usepackage{paralist} 

\usepackage[dvipsnames]{xcolor} 
\usepackage{changebar} 
\newcommand{\hint}[2]{} 

\newcommand{\FixedGap}[1][]{\ifthenelse{\equal{#1}{}}{\textsc{2-Gap}}{\ensuremath{\textsc{Gap}_{#1}}}\xspace}

\title{Evolutionary learning of fire fighting strategies}
\titlerunning{VC-dimension for $L_1$-visibility} 
\author{Martin Kretschmer
\and
Elmar Langetepe
}
\authorrunning{Kretschmer and Langetepe} 
\tocauthor{}
\institute{
University of Bonn, Department of Computer Science,
D-53113 Bonn, Germany}

\begin{document}

\maketitle
\begin{abstract}
The dynamic problem of enclosing an expanding fire can be modelled by a 
discrete variant in a grid graph. While the fire expands to all neighbouring cells 
in any time step, the fire fighter is allowed to block~$c$ cells in the average outside 
the fire in the same time interval. It was shown that the success of the fire fighter 
is guaranteed for $c>1.5$ but no strategy can enclose the fire for $c\leq 1.5$. 
For achieving such a critical threshold 
the correctness (sometimes even optimality) of strategies and lower bounds have been 
shown by integer programming or by direct but often very sophisticated arguments.  
We investigate the problem whether it is possible to find or to approach 
such a  threshold and/or optimal strategies by means of evolutionary algorithms, i.e., we just try to learn 
successful strategies for different constants~$c$ and have a look at the outcome. 
The main general idea is that this approach might give some insight in the power 
of evolutionary strategies for similar geometrically motivated threshold questions.  
We investigate the variant of protecting a highway with still unknown threshold 
and found interesting strategic paradigms. 
\end{abstract}

\section{Introduction}\label{intro-sect}

In the field of motion planning, online algorithms or Computational Geometry 
(and of course in many other areas) there are many examples of 
annoying gaps between upper and lower bounds of interesting  and important 
constants or running times. For establishing close bounds many theoretical attempts 
and different sophisticated approaches have been tried. 


A challenging approach might be to close or reduce such gaps (or even only get 
some more insight) by means of rather simple but efficient evolutionary or genetic approaches. If some structural properties or insight is known we can even apply more 
goal oriented 
algorithms. We would like to find out how far this might work. 
 Rather than analysing evolutionary algorithms theoretically as 
 for example given in \cite{bsw-haea-02,djw-aea-02}, we would like 
 to analyse the power of such simple algorithms for getting insight in well-defined 
 theoretical questions. Some challenging examples beyond the problems considered here are presented in Section~\ref{Future-sect}. 
 
 Overall our experimental approach can be seen as an Evolutionary Computation 
or an Evolutionary Algorithm for optimizing a population of solutions 
by natural selection  and mutation such that a fitness gradually increases; 
see~\cite{f-ectnp-95,h-anas-75,r-eotsp-73,s-nocm-81}. 
 
In this paper we concentrate on the context of discrete fire fighting in different 
variants. An overview of results in this context and some related problems
is given by Finbow \& MacGillivary \cite{finbow2009firefighter}.
Assume that in a grid-cell environment a cell that is on fire expands the 
fire from one cell to its four neighbouring cells in one time step.  
On the other hand the fire fighter can block some of the cells outside the 
fire in any time step. The number of cells that can be blocked is given 
by an asymptotic budget $c\geq 1$ such that at any time step $t$ 
we could have made use of $\lfloor c\times t\rfloor$ blocked cells in total.

We examine two questions. It is well-known that for $c>1.5$ an expanding fire 
can be enclosed; see \cite{ng2005fractional}. The result is  obtained by a sophisticated 
recursive strategy idea. Optimality (minimum number of burned cells) can be obtained
for example for $c=2$ by making use of ILP formulations; see \cite{wang2002fire}. This does not work 
well for smaller $c$ because of the running times.
On the other hand for $c\leq 1.5$ no strategy can stop the fire, 
shown by a tricky proof in~\cite{fh-fne-13}. Therefore $c=1.5$ is the fixed threshold
for this case. 

For this well-understood scenario we make use of 
simple evolutionary rules and show that for $c=2$ we 
obtain the optimal strategy extremly fast. For $c\geq 1.7$ we still obtain enclosement
results that seem to be close to the optimal. For $c$ less than $1.6$ our approach fails. 
The results are presented in Section~\ref{Enclosement-sect}. 

 The above first results might be seen as a test scenario for a 
new question considered in Section~\ref{Protection-sect}.
For a protection budget~$c$ the task is rather than enclosing the 
fire, we would like to prevent a highway from being reached 
by the fire soon. Theoretical results and a fixed threshold for this setting are still  unknown.  We try to get an impression 
how reasonable strategies look like for different values of bugdet $c<1.5$. 
It is more likely to make use of a single barrier close to the fire or is 
it recommendable to build (multiple) barriers away from the fire close to the highway? 
The focus here is that we get some ideas or insights by the use of evolutionary methods.
In contrary to the former enclosement problem we first make experiments and 
an ongoing task is to find formal proofs. 
The results and the corresponding conjectures are presented 
in Section~\ref{Protection-sect}.

The main conclusion of our work is that simple, goal oriented evolutionary strategies 
could help to give insight into the solutions of dynamic motion planning
problems. Especially, if such problems come along with a threshold question.  
The hope is that such approaches can also be used for similar problems. 
Some examples are given in Section~\ref{Future-sect}.

\section{Fire enclosement in a discrete grid settings}\label{Enclosement-sect}

Given an infinite grid graph with vertex set $\mathbb{Z}^{2}$. 
Each vertex represents a cell in a grid graph. In the following vertices and 
cells are handle as synonyms. 
The set of 
edges is given by $\left\{ \left(\left(u,v\right),\left(x,y\right)\right)\mid\left|u-x\right|+\left|v-y\right|=1\right\} $, i.e. each cell is neighbour to the cell directly above, below, left,
and right. A fire starts at $\left(0,0\right)$ and spreads over time.
After each time step, all cells with a burning neighbor start burning
as well. 

In the first setting the goal is to enclose the fire, such that only a finite (minimal)
number of cells is lost. To achieve this, a certain number of non
burning cells can be protected at each time step, which will then
never catch fire. 

\begin{figure}
\begin{centering}
\emph{Start} \includegraphics[width=2cm,page=1]{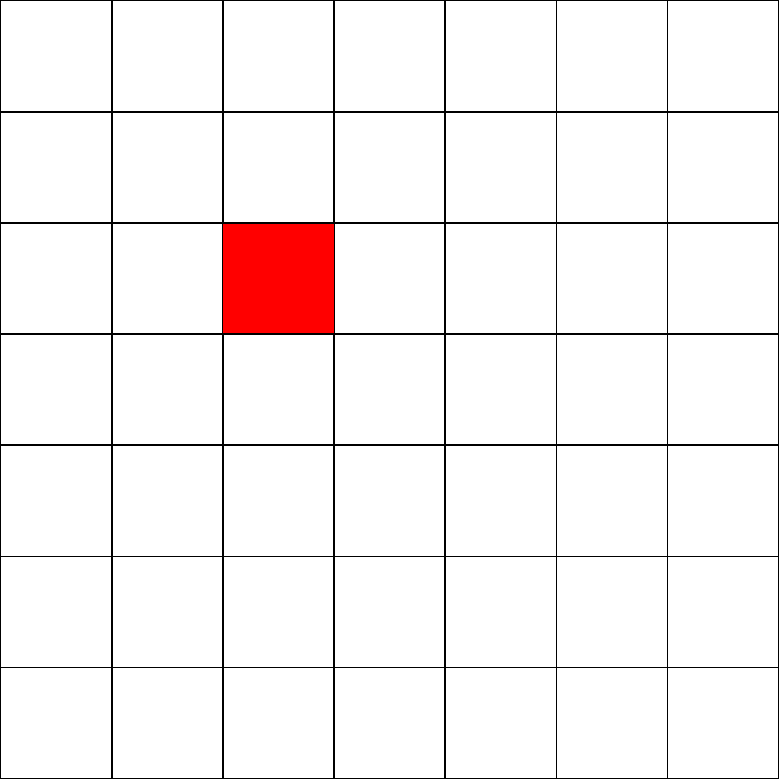}\hspace*{0.03cm}
$t=1$ \includegraphics[width=2cm,page=2]{FireExample12.pdf}\hspace*{0.03cm}
\emph{Spread} \includegraphics[width=2cm,page=3]{FireExample12.pdf}\hspace*{0.03cm}
$t=2$ \includegraphics[width=2cm,page=4]{FireExample12.pdf}
\vspace*{0.3cm}

\emph{Spread} \includegraphics[width=2cm,page=5]{FireExample12.pdf}\hspace*{0.03cm}
$t=3$ \includegraphics[width=2cm,page=6]{FireExample12.pdf}\hspace*{0.03cm}
\emph{Spread} \includegraphics[width=2cm,page=7]{FireExample12.pdf}\hspace*{0.03cm}
$t=4$ \includegraphics[width=2cm,page=8]{FireExample12.pdf}
\par\end{centering}
\caption{\label{SimpleExample-fig} An example for threshold $c=2.7$. The fire starts at a single cell. At any time step, 
the fighter blocks the remaining cells of its overall budget $\lfloor t\times 2.7\rfloor$  outside the fire. Then the fire spreads. 
The protected (black) cells are labelled by time parameters. 
After $4$ time steps the fire is enclosed. }
\end{figure}

The number of cells that can be blocked is given 
by an asymptotic threshold $c\geq 1$ such that at any time step $t$ 
we could have made use of $\lfloor c\times t\rfloor$ blocked cells. 
A simple example for $c=2.7$ is show in Figure~\ref{SimpleExample-fig}. 
In the first step the fire fighter blocks $\lfloor 1\times 2.7\rfloor=2$ two cells outside the fire. After the fire spreads 
in the next step 
the fire fighter blocks $\lfloor 2\times 2.7-2\rfloor=\lfloor 3.4\rfloor=3$ non-burning cells. 
Then the fire spreads again and in 
step 3 again $\lfloor 3\times 2.7-5\rfloor=\lfloor 3.1\rfloor=3$ cells can be blocked 
by the fire fighter outside the fire. The fire spreads for the last time 
and by blocking $\lfloor 4\times 2.7-8\rfloor=\lfloor 2.8 \rfloor=2$  cells in the fourth step
the fire is enclosed. 

 It has been shown that a fire can always be enclosed protecting $c=2$
cells at each time step and it is impossible to do so with only one
\cite{fogarty2003catching,wang2002fire}. Finally, it was proved that a fire can
always be enclosed when the average number~$c$ of protected cells 
exceeds 1.5 \cite{ng2005fractional}. This bound is tight as shown by \cite{fh-fne-13}. 

In the case of $c=2$ even an optimal
solution (i.e. minimal number of burning cells) has been found by using
Integer Linear Programming \cite{wang2002fire}. Compared to that, 
in the following we want
to investigate how good a simple evolutionary inspired algorithm can solve
this task and how close we can get to the thresholds. 
The first experiments  also can be seen as  a test scenario for the question 
of protecting a highway considered in Section~\ref{Protection-sect}.

\subsection{A goal oriented evolution model}\label{EnclosementEvolution-sect}

To use an evolutionary method, we require  a  formal description of a general
strategy, which can be modified (mutation) and recombined (inheritance)
to obtain a new strategy. Additionally, we have to define a fitness function
for the comparison of  strategies. Intuitively (and also driven by the known theoretical results) it seems to be  a 
good idea for a strategy to 
\begin{itemize}
\item start close to the fire
\item build a (more or less) connected chain of protected cells, trying  to surround the fire
\end{itemize}

\begin{figure}
\begin{centering}
\includegraphics[width=2.5cm,page=6]{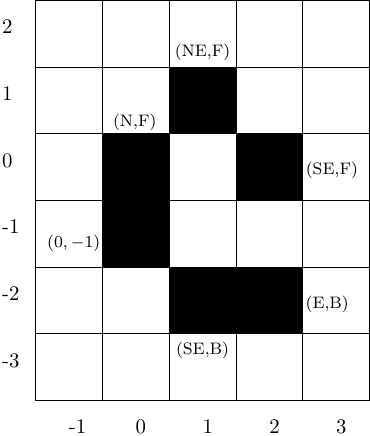}
\par\end{centering}
\caption{\label{fig:ExampleStrat}Example of a strategy starting at $\left(0,-1\right)$
with sequence $\left(\left(N,F\right),\left(NE,F\right),\left(SE,B\right),\left(SE,F\right),\left(E,B\right)\right)$.
Each protected vertex is labelled by the tuple that caused its 
protection.}
\end{figure}

We  further confirm these intuitions by having tried other variants as well. 
Our evolutionary experiments showed,  that strategies which start protecting vertices 
further away from the origin perform worse than strategies that start close to
the origin, some results for this are presented in Table~\ref{tab:different-starts}.
Analogously, the experiments showed that 
multiple disconnected barriers (that finally might be connected) do not work well. 
We omit to show the corresponding  experiments 
due to space constraints. We refer to Section~\ref{Protection-sect} where we have 
similar results for the problem of protecting a highway. 
The following definition is designed to follow the above simple principles.
\begin{definition}
A strategy consists of 
\begin{itemize}
\item a starting point $\left(i,j\right)$
\item a sequence of directions $\left\{ North,NorthEast,East,\ldots\right\} $
and each direction is combined with the information whether to extend the front ($F$) or
the back ($B$) of the chain
\end{itemize}
For short the strategy is given by the starting point and a list of pairs $(X,Y)$ with $X\in\{N,NE,E,\ldots\}$ and 
$Y\in\{F,B\}$. 
\end{definition}
An example of a strategy  (without a fire spread) is given in Figure \ref{fig:ExampleStrat}. For the fixed starting cell $(0,-1)$ the sequence $\left(\left(N,F\right),\left(NE,F\right),\left(SE,B\right),\left(SE,F\right),\left(E,B\right)\right)$  is applied 
as follows. By $\left(N,F\right)$ we extend  $(0,-1)$ \emph{forward} by the cell $(0,0)$  in the north which now is the \emph{new} front cell of the barrier. Then by $\left(NE,F\right)$ relative to 
the new front cell we block the cell in direction north-east, which is cell $(1,1)$. 
After that we apply $\left(SE,B\right)$ for the current back end of the barrier which still is $(0,-1)$. The new back end cell is $(1,-2)$ which lies south-east from $(0,-1)$ and so on. 

Notice, that such a strategy does not contain the information of the
time at which the next vertex is protected. Instead, the next tuple
of the sequence is applied, whenever we are allowed to protect an
additional vertex.

The number of vertices that are protected per step is based on a
bank account idea. We start with an initial budget and each time a
vertex is protected, the budget decreases by 1. The budget has to remain 
positive but is always fully exhausted. 
After the fire has
spread by one step, the budget increases by the fixed amount $c$. E.g.
$c=2$ means we can protect exactly two vertices in any step. For
$c=1.5$, the number of protected vertices alters between 1 and 2.

\newpage

\subsubsection{Handling illegal genomes}

The above genome design still allows for ``illegal'' genomes. This means that 
a strategy tries to protect cells which are already burning. 
For example in Figure \ref{fig:ExampleStrat}, if the fire starts in cell $(0,0)$ 
the first pair $\left(N,F\right)$ of the given sequence cannot be applied. 

One could simply stop the
protection of any further vertices in this case. This will result in an uncontrolled
spread of the fire and thus has a bad fitness. 
Especially for randomly initialized strategies this could happen very early
and is not recommendable.  

Another approach would be to simply ignore the protection of burning
cells and skip to the next element of the strategy sequence. 
 In this case, it might happen that we skip through the whole sequence
very quickly.

To avoid the above drawbacks we decided to use the following behaviour. 
Whenever the
sequence tries to protect a cell that is already burning, we start
a search for the next non-burning cell in clockwise or counter clockwise
order, depending whether we want to extend the front or the back of
the barrier, starting at the direction that is given by the sequence.
For example in Figure~\ref{fig:ExampleStrat}, if $(0,0)$ is burning in the 
beginning, the application of $\left(N,F\right)$ from $(0,-1)$ results in 
blocking the cell $(-1,1)$, which gives the new front. 

\subsubsection{Fitness Evaluation}

In order to determine the fitness of a strategy 
two values seem to be important.  The time  needed to enclose 
the fire and the total number of burning vertices. Since randomly initialized
sequences will most likely not enclose the fire, we use the total
number of burning vertices after a fixed simulation time~$t$. This also gives rise to gradual improvements. For example in Figure~\ref{SimpleExample-fig} 
for a simulation time $t=3$ the given strategy 
has fitness $5$, since $5$ cells are burning at time $t=3$. 
Note that we run arbitrary strategies with different simulation times (or steps). 

\subsection{Evolutionary Algorithm}\label{EnclosementAlg-sect}

The following algorithm keeps improving a randomly initialized set
of strategies until it is manually stopped. Besides the budged $c$,
it has several parameters which determine its behaviour.
\begin{itemize}
\item Input:
\begin{itemize}
\item $c$ budget income per time step
\item $n$ population size
\item $t$ number of simulation steps
\item $p$ mutation probability
\item $r$ ratio of parents kept after external selection
\end{itemize}
\item Initialization: A population $P$ of $n$ randomly generated strategies
(except the start point which is fixed to $\left(0,1\right)$), each
strategy needs to have a sequence of length at least $t\cdot c$
\item repeat
\begin{itemize}
\item simulate any strategy of $P$ for $t$ time steps and determine its
fitness
\item order $P$ by fitness in increasing order and keep only the best
$\left\lfloor r\cdot n\right\rfloor $ strategies as parents
\item restock $P$ again to size $n$ by selecting two parent strategies
and combining their sequences via single-point crossover
\item for each tuple in each sequence of $P$, change it with probability
$p$ to a new random direction and extension side
\end{itemize}
\end{itemize}
Note that for speeding up the results of our simulation as 
 presented in the next section for $c<2$ we decided to start the algorithm with an 
initial budget of $2$. This allows us to protected two cell in the first step. Our
experiments showed that this allows our algorithm to find successful
strategies much faster and therefore also for smaller values of~$c$.
Asymptotically, there is no difference for the threshold. 
This small artefact might also be interpreted as a goal oriented approach. 

\subsection{Experimental results}\label{EnclosementResults-sect}
\begin{figure}
\begin{centering}
\begin{minipage}[t]{0.8\columnwidth}%
\fbox{\begin{minipage}[t]{0.33\columnwidth}%
\begin{center}
\includegraphics[width=0.5\textwidth]{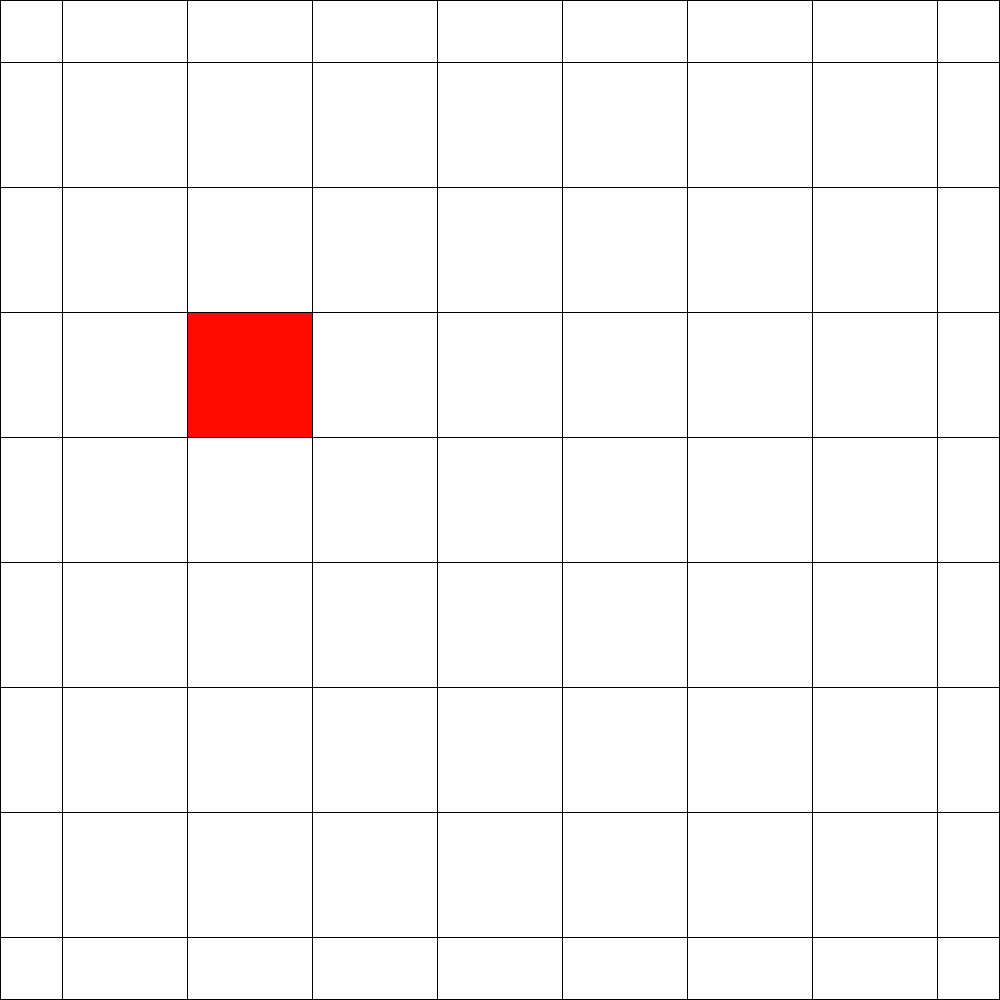}
\par\end{center}%
\end{minipage}}%
\fbox{\begin{minipage}[t]{0.33\columnwidth}%
\begin{center}
\includegraphics[width=0.5\textwidth]{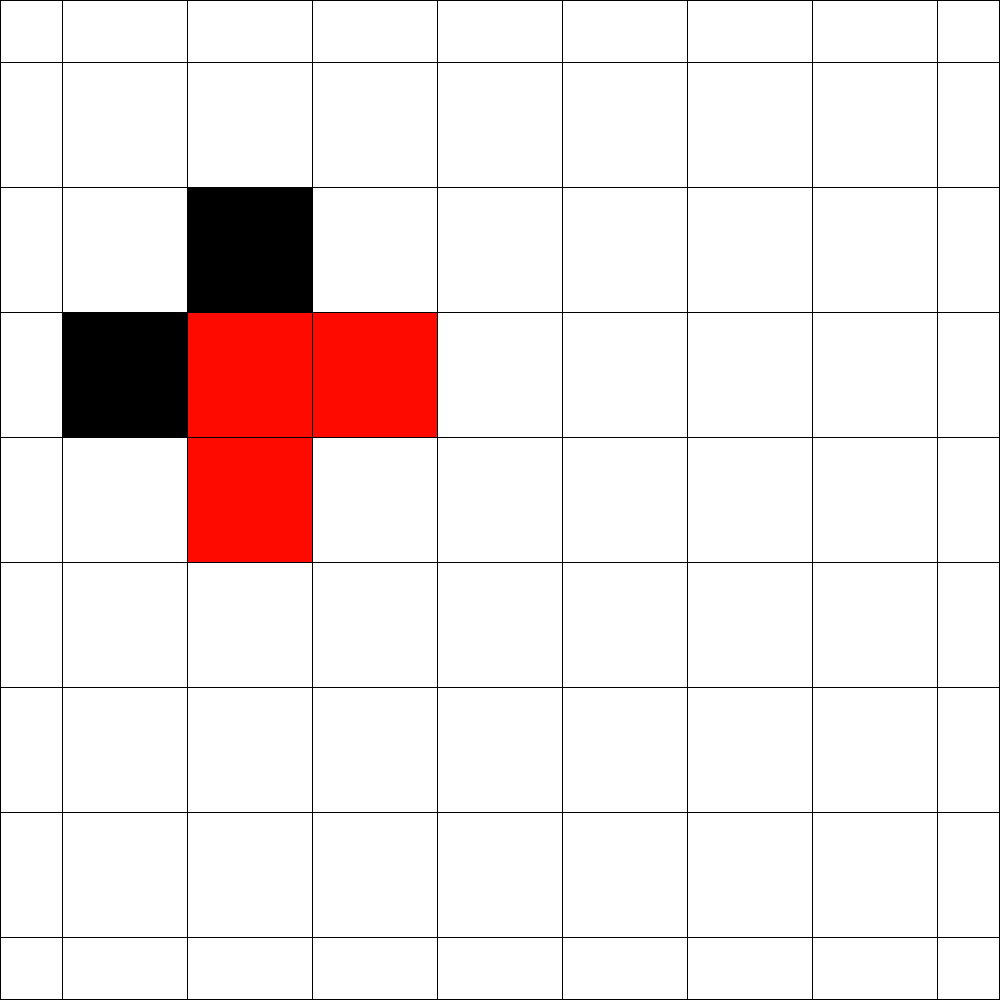}
\par\end{center}%
\end{minipage}}%
\fbox{\begin{minipage}[t]{0.33\columnwidth}%
\begin{center}
\includegraphics[width=0.5\textwidth]{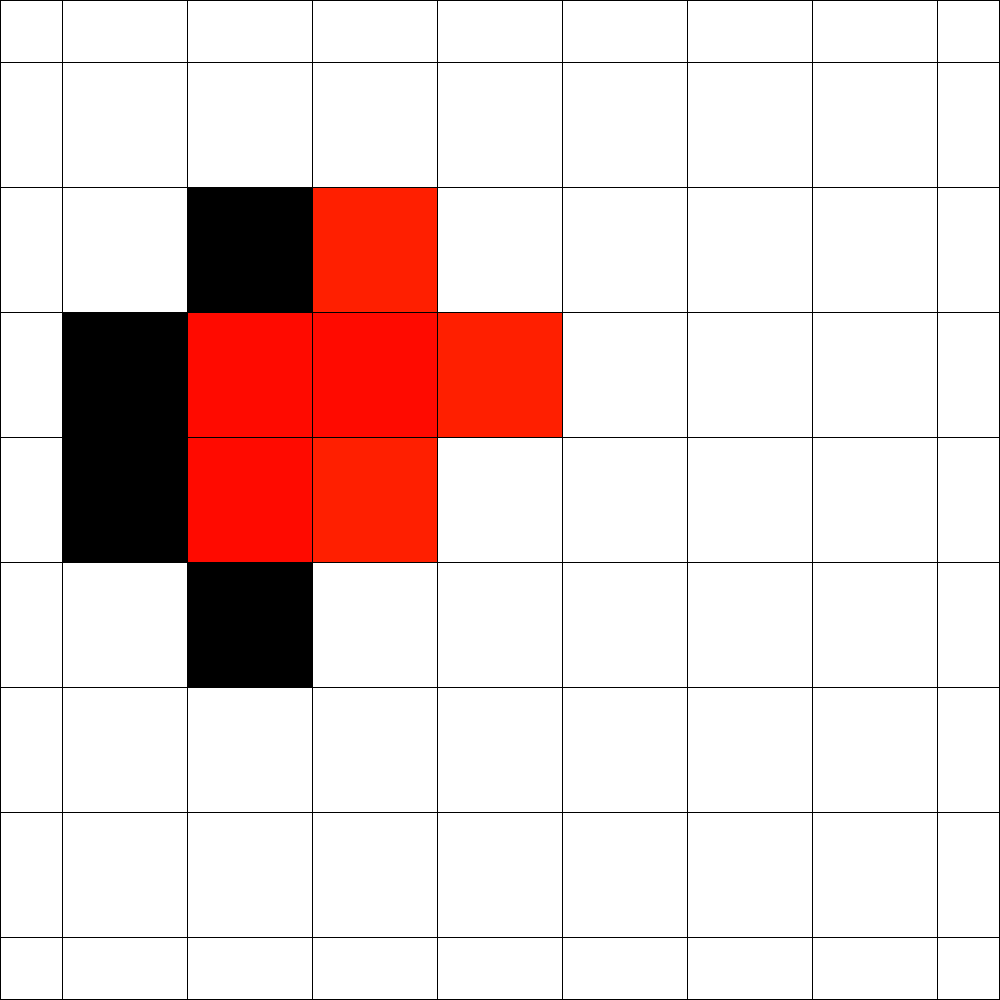}
\par\end{center}%
\end{minipage}}\\
\fbox{\begin{minipage}[t]{0.33\columnwidth}%
\begin{center}
\includegraphics[width=0.5\textwidth]{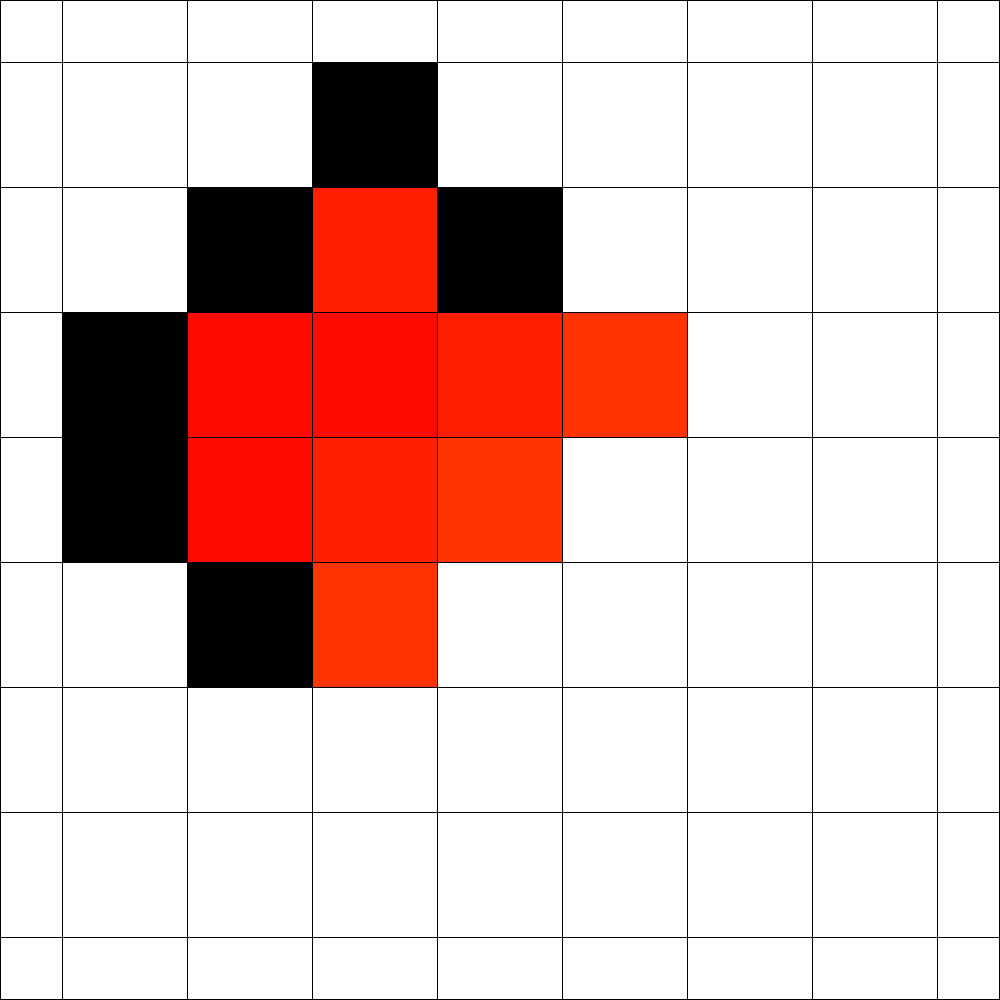}
\par\end{center}%
\end{minipage}}%
\fbox{\begin{minipage}[t]{0.33\columnwidth}%
\begin{center}
\includegraphics[width=0.5\textwidth]{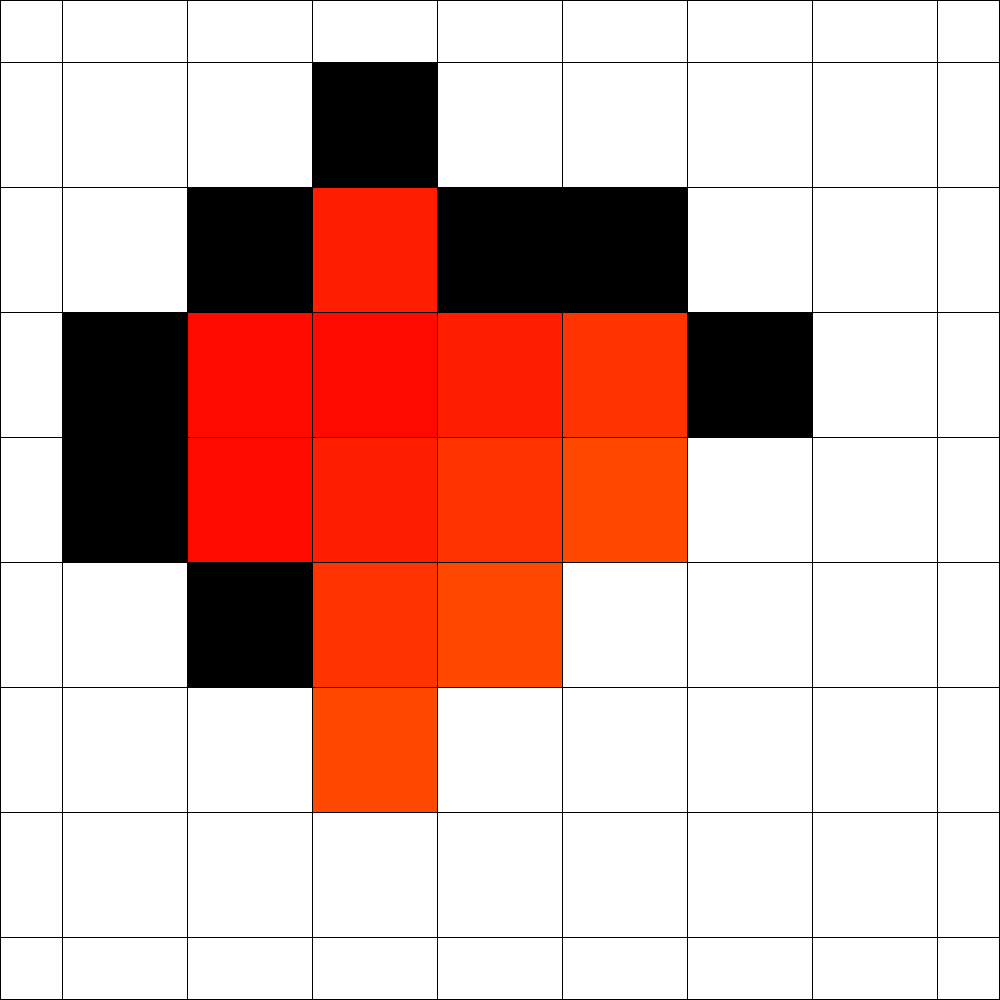}
\par\end{center}%
\end{minipage}}%
\fbox{\begin{minipage}[t]{0.33\columnwidth}%
\begin{center}
\includegraphics[width=0.5\textwidth]{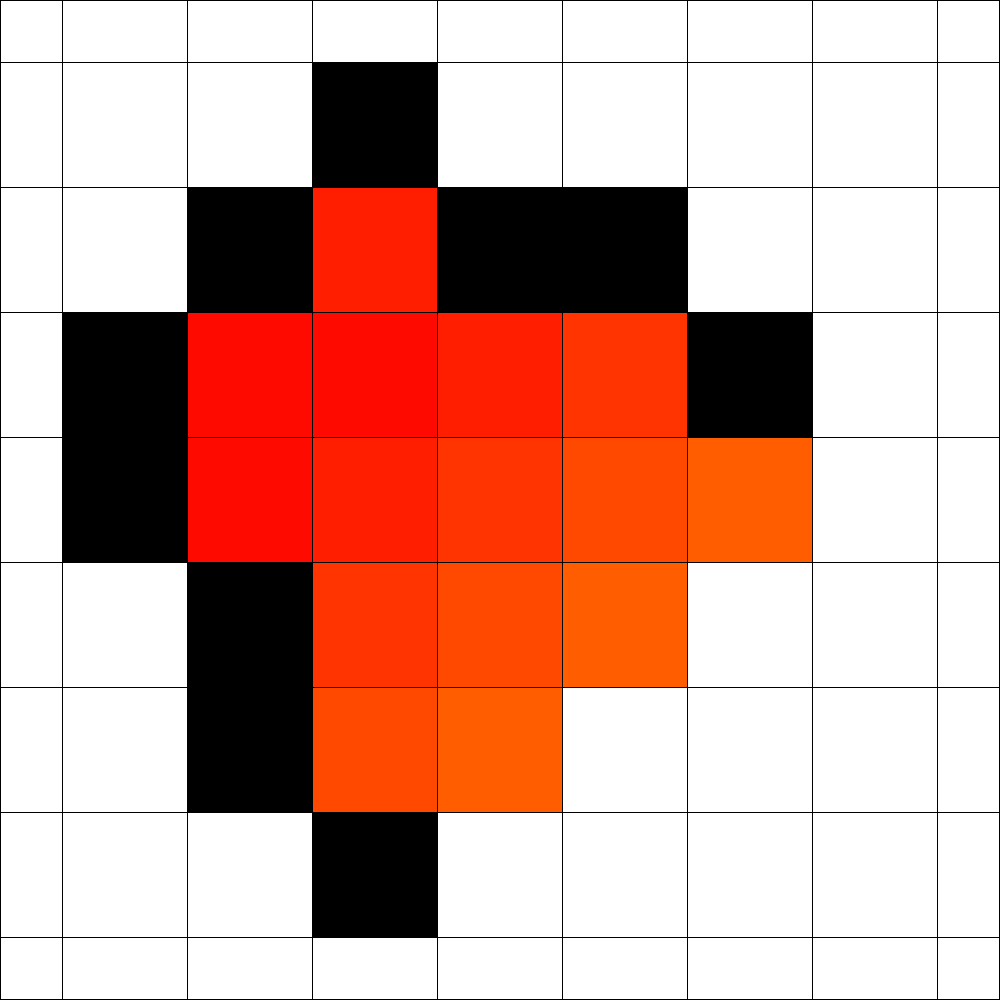}
\par\end{center}%
\end{minipage}}\\
\fbox{\begin{minipage}[t]{0.33\columnwidth}%
\begin{center}
\includegraphics[width=0.5\textwidth]{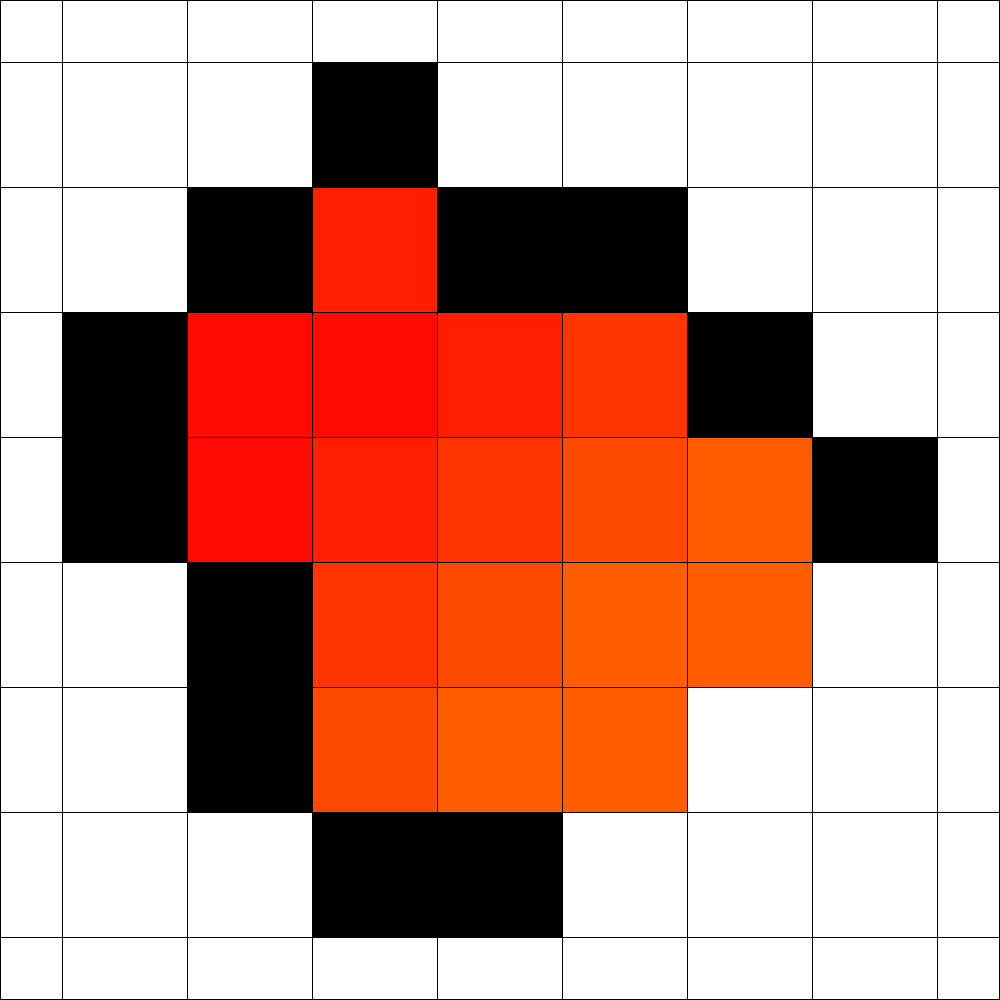}
\par\end{center}%
\end{minipage}}%
\fbox{\begin{minipage}[t]{0.33\columnwidth}%
\begin{center}
\includegraphics[width=0.5\textwidth]{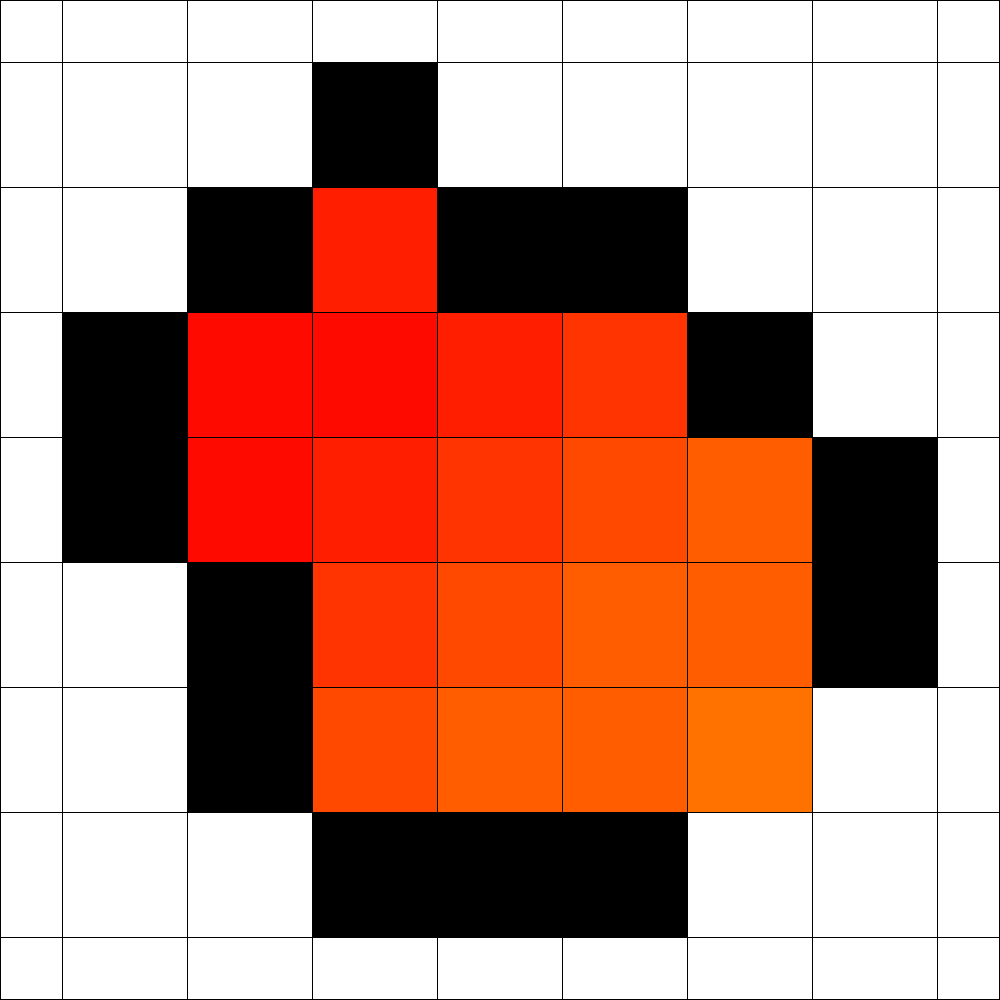}
\par\end{center}%
\end{minipage}}%
\fbox{\begin{minipage}[t]{0.33\columnwidth}%
\begin{center}
\includegraphics[width=0.5\textwidth]{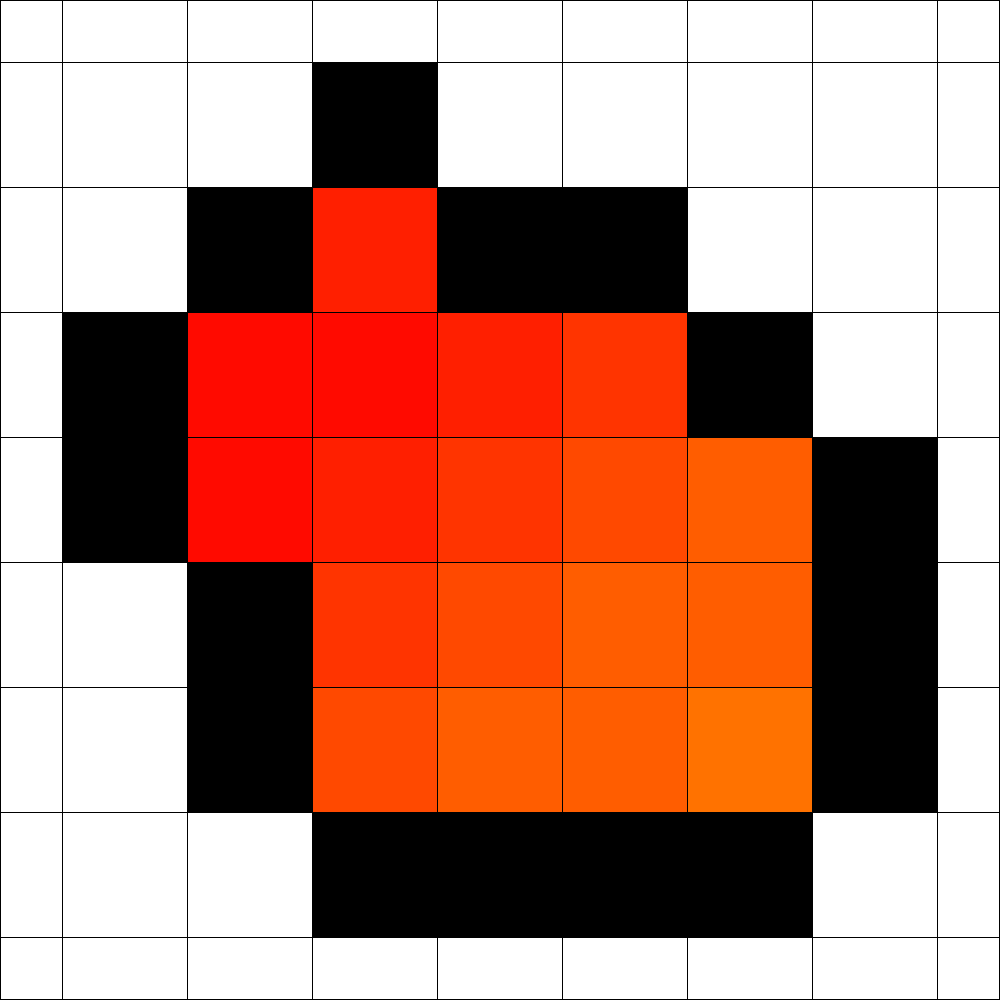}
\par\end{center}%
\end{minipage}}
%
\end{minipage}
\par\end{centering}
\caption{\label{fig:Strat2}Example of a strategy found when protecting exactly
$c=2$ vertices per step. Enclosed after 8 steps with 18 burning vertices.}
\end{figure}
Figure \ref{fig:Strat2} shows an optimal  strategy that was found  by evolution for the case $c=2$. It takes 8 steps to enclose the 
fire and in the end 18 vertices are on fire. This is optimal for both time and
number of burning vertices as shown in \cite{fogarty2003catching}. 
Surprisingly, it tooks only 84 generations in total until this strategy was found.

\begin{figure}
\begin{centering}
\begin{minipage}[t]{0.75\columnwidth}%
\begin{center}
\fbox{\begin{minipage}[t]{0.45\columnwidth}%
\begin{center}
\includegraphics[width=0.75\textwidth]{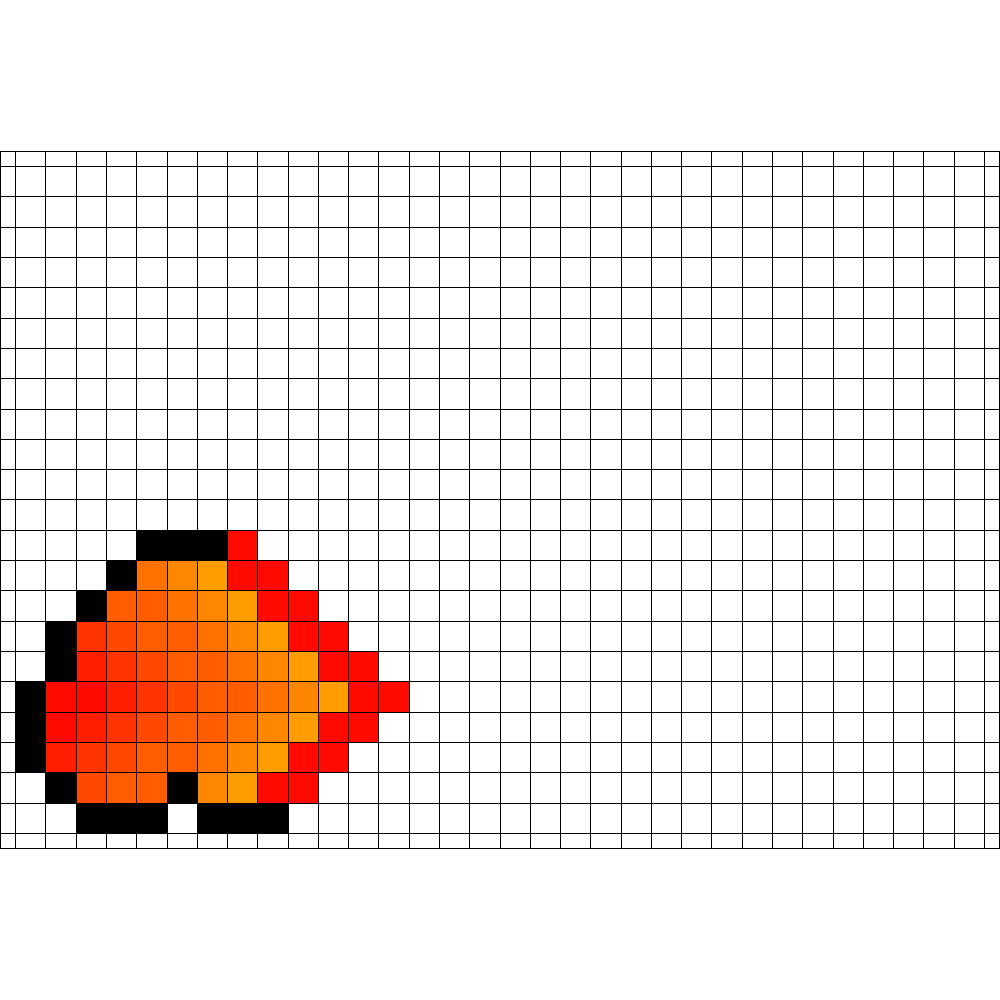}
\par\end{center}%
\end{minipage}}%
\fbox{\begin{minipage}[t]{0.45\columnwidth}%
\begin{center}
\includegraphics[width=0.75\textwidth]{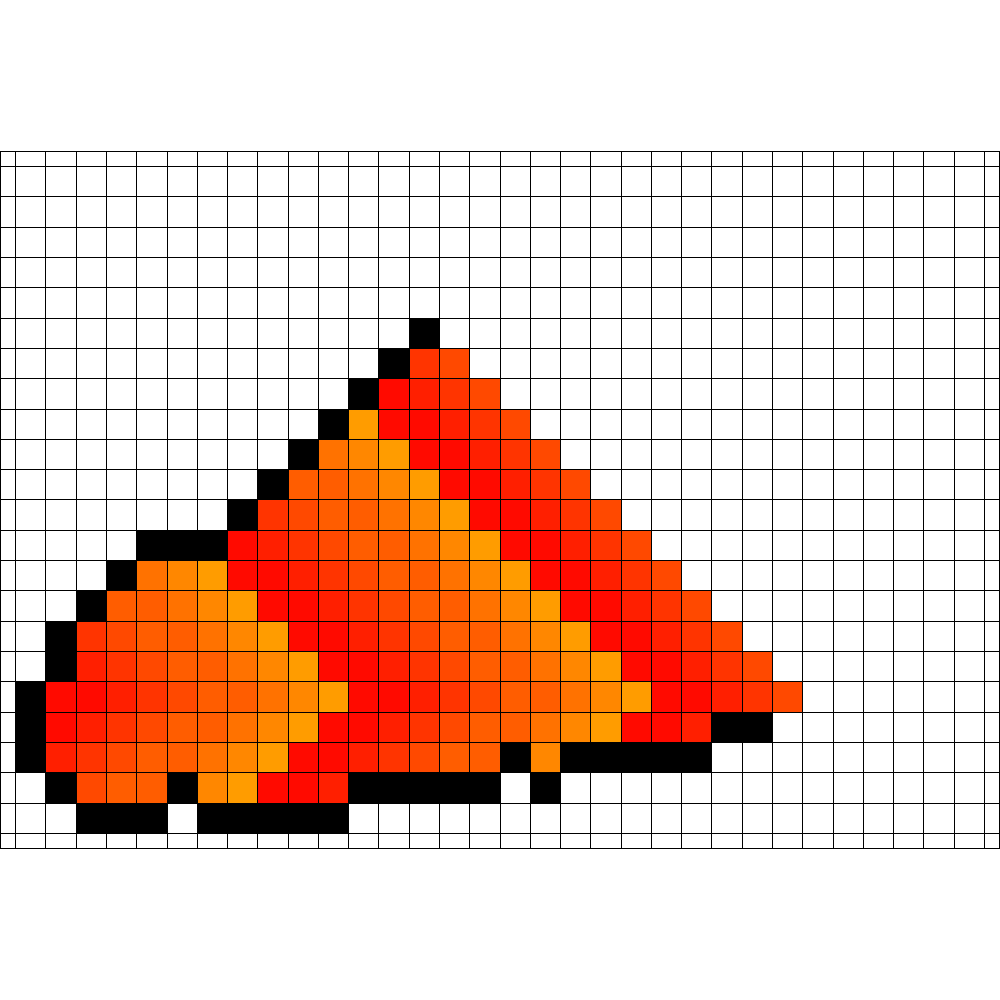}
\par\end{center}%
\end{minipage}}\\
\fbox{\begin{minipage}[t]{0.45\columnwidth}%
\begin{center}
\includegraphics[width=0.75\textwidth]{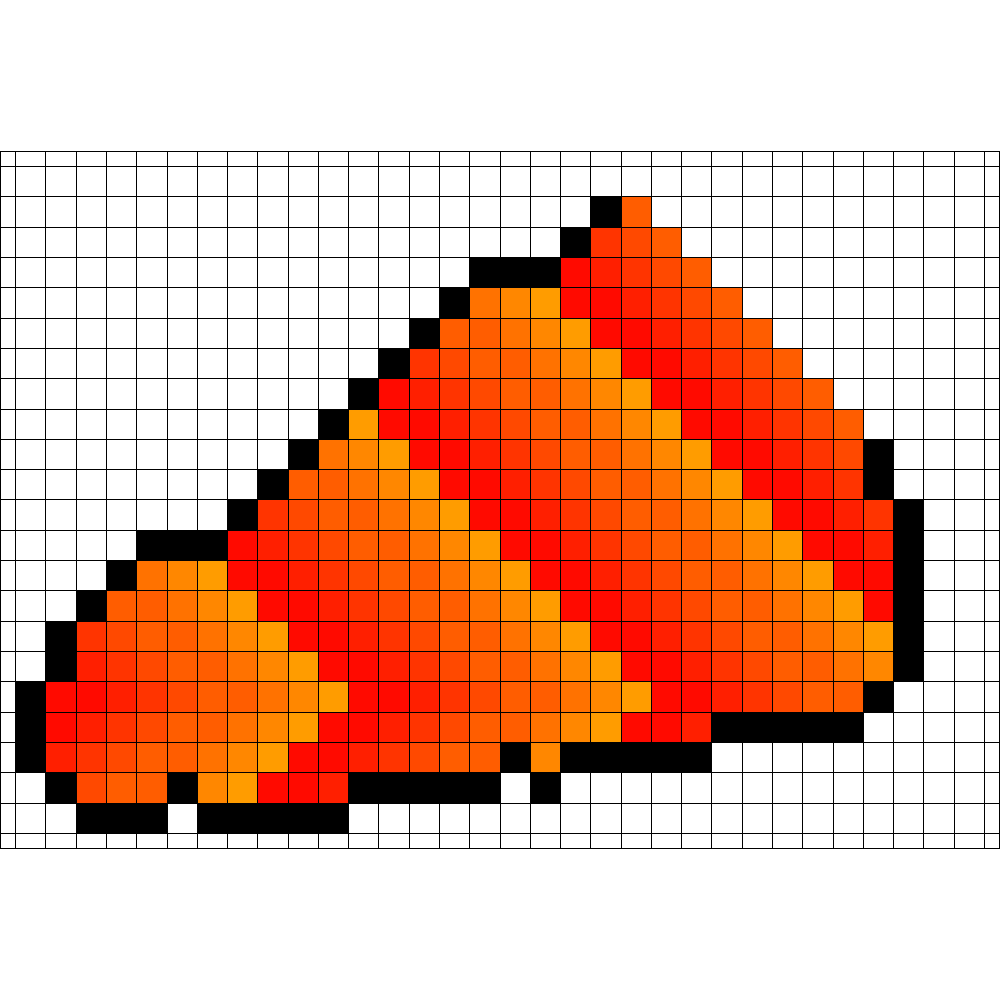}
\par\end{center}%
\end{minipage}}%
\fbox{\begin{minipage}[t]{0.45\columnwidth}%
\begin{center}
\includegraphics[width=0.75\textwidth]{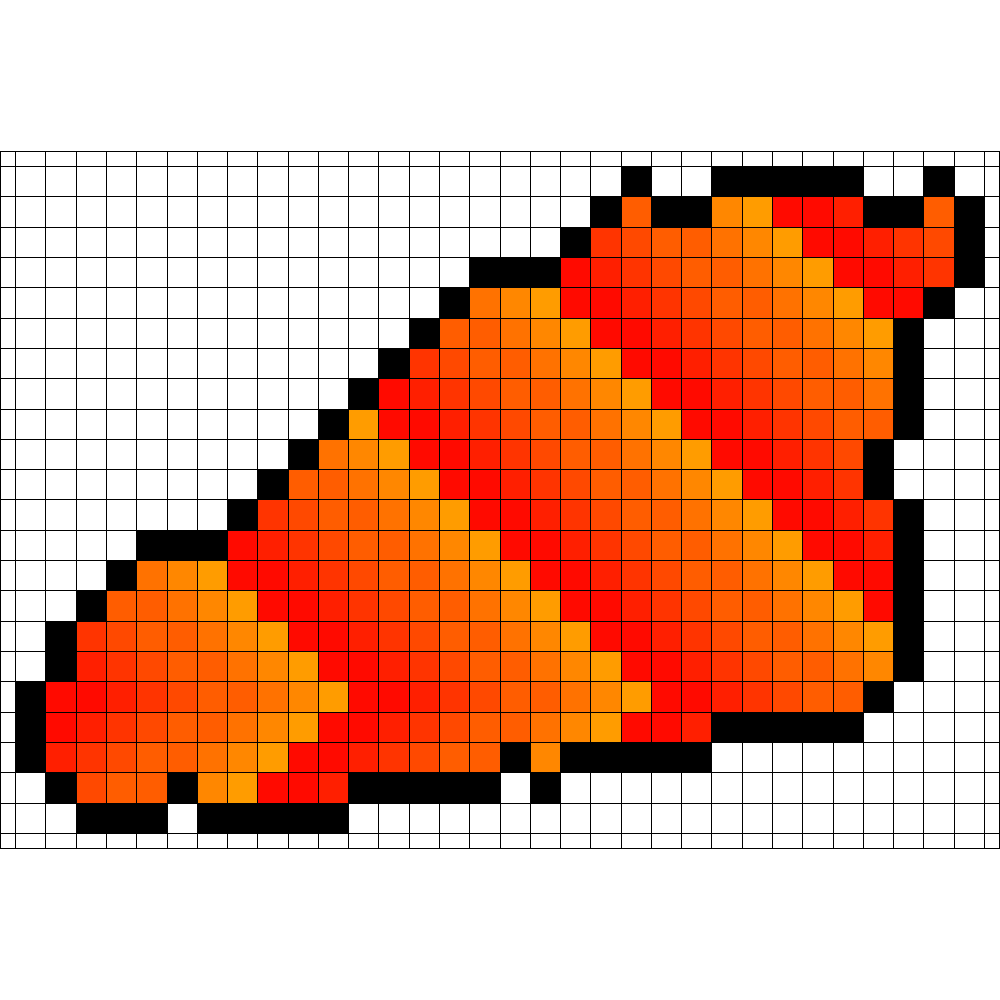}
\par\end{center}%
\end{minipage}}
\par\end{center}
%
\end{minipage}

\par\end{centering}
\caption{\label{fig:Strat1.8} Strategy found for $c=1.7$. Enclosed after 46
steps with 371 burning vertices. The colored shading indicates how the fire
spreads over time. }
\end{figure}

An example of a strategy that was found for $c=1.7$ is depicted in
Figure \ref{fig:Strat1.8}. 
Finding this strategy took
1002 generations in total. A video of a successful strategy is shown in\\
 \hspace*{1cm} \url{http://tizian.informatik.uni-bonn.de/Video/1.7Enclosing.mp4} .
 
\begin{figure}
\begin{centering}
\includegraphics[width=0.45\textwidth]{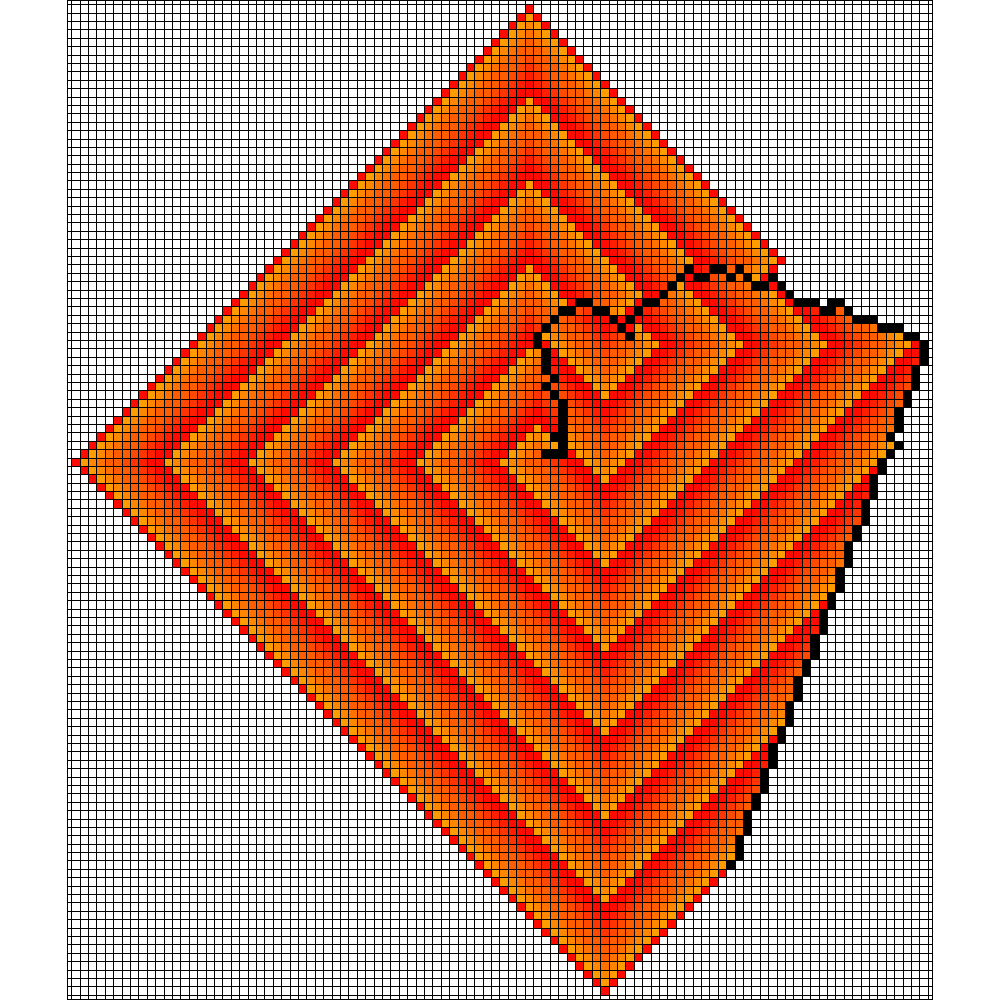}
\par\end{centering}
\caption{\label{fig:Failed-strategies} Failed strategy for $c=1.6$. 
Note that the fire expands on both sides of the barrier.}
\end{figure}

For even smaller values of $c$, our algorithm starts failing to find
enclosing strategies. An example for $c=1.6$ is given in Figure
\ref{fig:Failed-strategies}. It seems that the strategy might
be able to enclose the fire after a longer time, but even increasing the
simulation time $t$ did not lead to success. 

\begin{figure}
\begin{centering}
\includegraphics[width=0.8\textwidth]{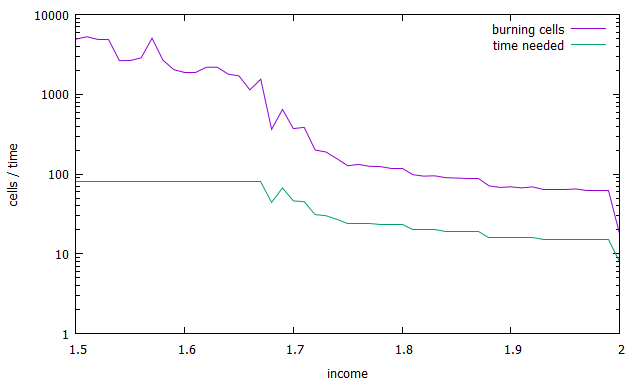}
\par\end{centering}
\caption{\label{fig:bench} For simulation time $t=80$ we obtain 
positive results up to $c$ slightly larger than $1.68$.}
\end{figure}

Figure \ref{fig:bench} shows for which values of $c$ we were able
to find enclosing strategies. We chose a simulation time of $t=80$
and one can see that for values smaller than $c=1.68$ the building
of the barrier continued until the simulation ended. This means that the fire
was not enclosed.

\begin{table}
\begin{centering}
\begin{tabular}{|c|c|c|}
\hline 
Start & Enclosing Time & Burning Vertices\tabularnewline
\hline 
\hline 
$\left(0,1\right)$ & 8 & 18\tabularnewline
\hline 
$\left(0,4\right)$ & 23 & 156\tabularnewline
\hline 
$\left(1,3\right)$ & 15 & 68\tabularnewline
\hline 
$\left(2,2\right)$ & 24 & 161\tabularnewline
\hline
\end{tabular}\\[2ex]
\par\end{centering}
\caption{\label{tab:different-starts} Fitness of best strategies found for
$c=2$ and different starting points.}
\end{table}

So far, any strategy presented had a fixed start point neighboring
the origin of the fire. As an example we compare the strategy for
$c=2$ mentioned above to strategies whose start point is fixed to
a vertex four steps away from the origin. Up to symmetry there
are three different coordinates for this. $\left(0,4\right)$, $\left(1,3\right)$
and $\left(2,2\right)$. Table \ref{tab:different-starts} shows the
times required to enclose the fire using these different starting points,
compared to the optimal strategy shown above.  Starting
further away from the fire takes longer to enclose the fire.
We have similar results for other values of~$c$.

\subsection{Fire enclosement conclusion}\label{EnclosementConcl-sect}

At least for values  of~$c$ a bit away from the overall tight threshold, 
the simple evolutionary goal oriented algorithm was able
to find successful (and in the case of $c\geq 2$ even optimal)
strategies surprisingly fast. Successful strategies close to the threshold 
$c=1.5$ are not easy to find. Actually this seems to be clear. The corresponding 
solution came along with some very sophisticated recursive strategies 
not easy to find by an evolutionary approach.

\section{Protection of a highway}\label{Protection-sect}

\newcommand{\currwidth}{5cm}
\begin{figure}
\begin{centering}
Start \includegraphics[width=\currwidth,page=1]{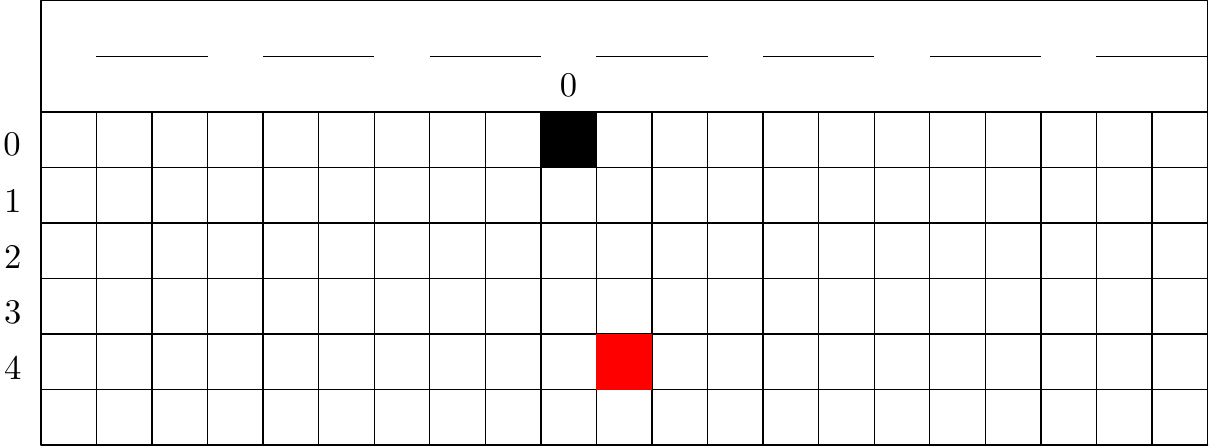}\hspace*{0.03cm}
$t=1$\includegraphics[width=\currwidth,page=3]{ProtectionExample.pdf}

\vspace*{0.3cm}
$t=2$\includegraphics[width=\currwidth,page=4]{ProtectionExample.pdf}\hspace*{0.03cm}
$t=3$\includegraphics[width=\currwidth,page=5]{ProtectionExample.pdf}

\vspace*{0.3cm}
$t=4$\includegraphics[width=\currwidth,page=6]{ProtectionExample.pdf}\hspace*{0.03cm}
$t=5$\includegraphics[width=\currwidth,page=7]{ProtectionExample.pdf}

\vspace*{0.3cm}
$t=6$\includegraphics[width=\currwidth,page=8]{ProtectionExample.pdf}\hspace*{0.03cm}
$t=7$\includegraphics[width=\currwidth,page=9]{ProtectionExample.pdf}
%
\par\end{centering}
\caption{\label{ExampleHighway-fig} A fire fighter strategy for protecting a highway 
with budget $c=1.2$. The fire starts at a single cell, one cell is initially protected.
At any time step, 
the fighter first blocks the remaining cells of its overall  $\lfloor t\times c\rfloor$ budget outside the fire and then the fire spreads. After $7$ time steps the fire reaches the highway.} 
\end{figure}

Here we consider a different and new question. Conversely to the previous section we did not have any idea for a reasonable strategy and/or a threshold. 
The question is how long can we protect a highway (modeled by a line of cells) 
from the fire, if some budget $c<1.5$ is given. We would like to avoid that the fire touches a \emph{line} very early? 
What is a reasonable strategy? 
Should we start close to the fire or close to the highway? 
Should we design a single connected barrier or more barriers which are
partly disconnected? 
  
In  Figure~\ref{ExampleHighway-fig} we give an example for a strategy for~$c=1.2$. This means that in the first 4 time steps the fire fighter makes use of a 
single blocking cell. In  step $t=5$ the
fire fighter can block two cells for the first time since $\lfloor 5\times 1.2-4\rfloor=2$ holds. 
Similar to the previous section we can also 
assume that in the start situation some constant cells are already blocked, 
this is indicated by the blocked cell of label $0$ in Figure~\ref{ExampleHighway-fig}. 
 Figure~\ref{ExampleHighway-fig} has to be interpreted as follows. If $c_{t-1}$ cells  were used from the budget 
of  the fire fighter after step $t-1$, at the next time step $t$, 
the fighter first blocks $\lfloor t\times c-c_{t-1}\rfloor$ cells outside the fire 
and then the fire spreads. After $7$ time steps and the corresponding spread 
the fire reaches the highway. 
Note that the strategy stops in this moment.

\subsection{Evolution models}\label{ProtectionEvolutionModel-sect}

Since the given problem was not theoretically analysed before, we
first had to test several ideas experimentally in order to achieve
a more goal oriented model.  In contrary to the enclosement 
scenario discussed before we do not know whether a connected
strategy  will lead to optimal or efficient solutions.
So we first tried to allow general strategies that could protect
arbitrary cells. We made use of a very simple coordinate based genome model,
such that a strategy is simply defined by a set of cell coordinates 
defining which cells should be protected. 

For such a set of cells, the cells are protected in their 
$L_{1}-$distance  order from the origin of the fire, i.e., cells closer to the 
fire origin will be protected first. In the evolution process this behaviour 
forces that useless protections far away from the origin will be cancelled out 
more quickly. In principle the above principles allow us to define arbitrary 
strategies. 

Altogether, we either make use of 
\begin{itemize}
\item a connected genome (as in the previous section) or
\item a coordinate genome described by a set of cell coordinates (as just mentioned)
\end{itemize}

\subsection{Evolutionary Algorithm}\label{ProtectionAlg-sect}

We noticed that usually we do not get any improvements by the recombination 
of strategies. Therefore we adapted the framework used in Section~\ref{EnclosementEvolution-sect} and restrict the algorithm to mutation only. This also means that we do
not need to have a large population, instead we only initialize
a single randomly generated strategy that will keep mutating. If a
mutation leads to an improvement, the strategy keeps that mutation,
otherwise it is undone.

This process of improving a single strategy can easily be parallelized 
such that a larger set of single strategies keeps improving over time. This
is very beneficial, because the final result often depends on the initialization
and not every run leads to the best result. 
Another difference to the previous section is the fitness evaluation,
which obviously has to be adjusted with respect to the  problem definition.

\subsubsection{Fitness evaluation}\label{ProtectionFitness-sect}

For the enclosement problem considered in Section~\ref{EnclosementEvolution-sect}
we tried to minimize the total number
of burning cells. In this case we  have used exactly this number for 
determining the
fitness. Now we want to maximize the time the fire requires to reach
the highway.  It turns out that increasing this time value directly 
 by a random mutation or recombination is very unlikely. 
Therefore we require  a fitness evaluation that also allows for smaller and  gradual improvements. To attain this we take into  account how many vertices 
are burning and also their corresponding distance to the fire. 
A formal definition is given  below.

For letting the algorithm run, actually we only need to be able to
compare strategies pairwise. Fortunately, this can also be realized by 
our fitness function. 

\begin{definition}
Let $S$ be a protection strategy  for a given highway.  By $r(S)$ we denote the first moment in time when the fire reaches the highway, if $S$ is applied. By $d(S)_i$ we denote the number of burning cells 
with distance $i$ to the highway after $r(S)$ simulation steps. 

Strategy $S_{1}$ has a \emph{larger fitness} than $S_{2}$ if 
$r(S_1)>r(S_2)$ holds or for $r(S_1)=r(S_2)$ if $d(S_1)_{i}<d(S_2)_{i}$ holds for the smallest index $i$ where $d(S_1)_{i}\neq d(S_2)_{i}$.
\end{definition}
For example in Figure~\ref{ExampleHighway-fig}  the given strategy $S$ has value $r(S)=7$. 
We also have $d(S)_0=1$, $d(S)_1=9$ and $d(S,9)_2=12$ and so on. 
So  another strategy $S'$ would have  larger fitness, 
if $r(S')>7$ holds or for  $r(S')=7=r(S)$, if we have for example 
$d(S')_0=1$, $d(S')_1=9$ and $d(S')_2=11<12=d(S)_2$. 

The main idea is that by trying to keep the fire farther away from the highway, 
finally also the overall time where the fire reaches the highway can be increased. 

\subsection{Experimental results}\label{ProtectionResults-sect}

Similar to the enclosement problem our implementation allows us to set or manipulate many different parameters and options for a goal oriented evolutionary process, such as the budget~$c$, the strategy design  (general genome or connected barriers), the population size (number of strategies optimized in parallel), 
the mutation rate, the fire source  (distance to the highway), 
starting positions (for connected barriers), optional initial budget  and so on. 

\subsubsection{Videos:} Finally and interestingly we mainly found two different strategic behaviours depending 
on the corresponding genomes, they will be explained precisely below. 
For convenience for $c=1.2$ we prepared two animations that show the 
finally attained best strategies for 

\begin{description}
\item[1. General genomes:] Symmetric and alternating strategy:\\
\url{http://tizian.informatik.uni-bonn.de/Video/1.2SymAlt.mp4}
\item[2. Connected barriers:] Asymmetric and diagonal strategy:\\
\url{http://tizian.informatik.uni-bonn.de/Video/1.2AsymDiag.mp4}
\end{description}
where in the second case of connected barriers sometimes 
also symmetric and alternating strategies were attained  under 
circumstances explained below. The above strategies have been found after 
156925 (1.)  and 34226 (2.) generations.

\begin{figure}
A) \includegraphics[width=0.45\textwidth]{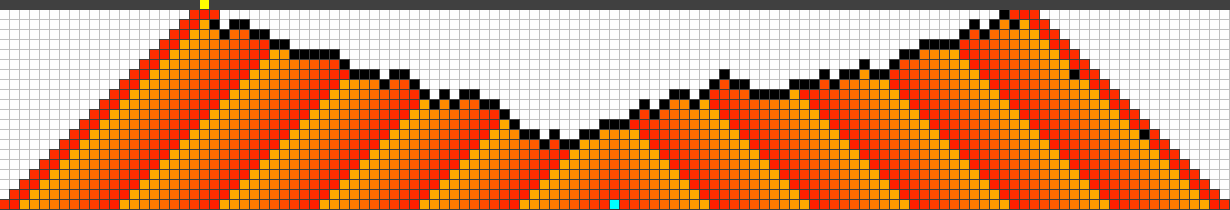}\hspace*{0.5cm}
B) \includegraphics[width=0.45\textwidth]{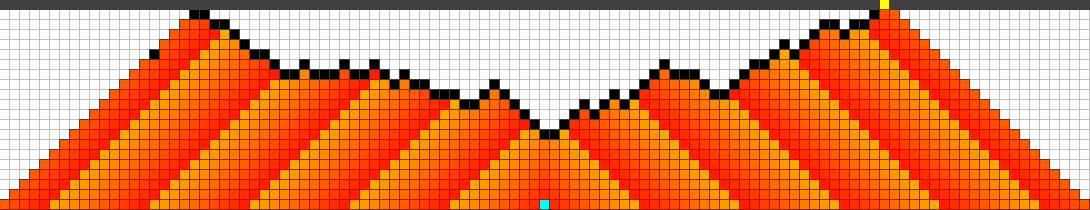}\\[3ex]
C) \includegraphics[width=0.45\textwidth]{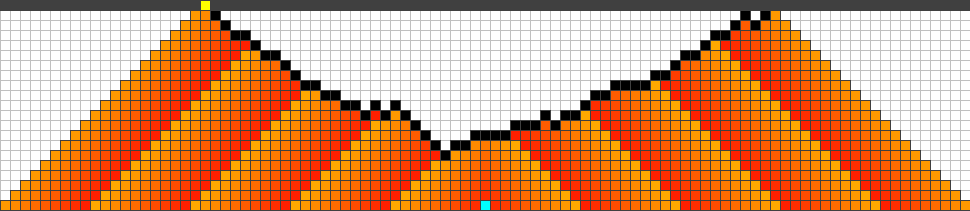}\hspace*{0.5cm}
D) \includegraphics[width=0.45\textwidth]{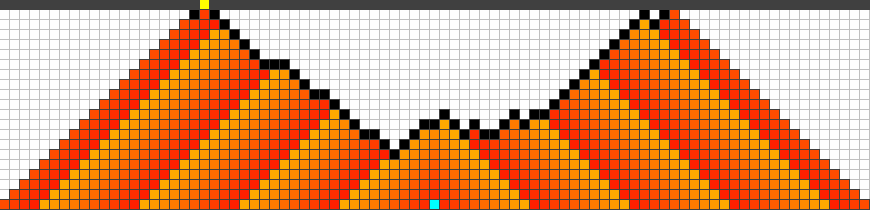}
\caption{\label{fig:different-scattered}Resulting strategies using the general coordinate
genome for different values of the budget: A) $c=1.4$,  B) $c=1.3$,  C) $c=1.2$ 
D) $c=1.1$. The
fire starts 20 steps away from the highway. The highway was reached
after 61, 54, 48 and 43 steps, respectively.}
\end{figure}

\paragraph{\bf 1. General genomes}
First, we found out that the use of general genomes always (for different 
settings) mutate toward  connected barriers; Figure~\ref{fig:different-scattered}
shows some of the finally  attained strategies. 
All strategies show a similar behaviour. They start somewhere between
the origin of the fire and the highway, usually a bit closer to the fire.
Then any strategy continues to protect cells alternating between left and right,
trying to keep the fire as long and as far away from the highway as possible.
In the following we refer to such strategies as \emph{symmetric and alternating}.

 Note that any of the given strategies can be reconstructed such that the 
symmetric and alternating process is performed directly at the highway. The time where the fire reaches the boundary will not change in this case. 
Our fitness function simply prefers to shift the fire away from the highway. 

\subsubsection{2. Connected barriers}\label{connectedgenome-prot-sect}
After that we again considered connected barriers 
with different starting 
positions below the origin. Depending on the distance between the start and the fire, we observed 
two different strategic behaviours which can be categorized as follows. 

\begin{figure}
\begin{centering}
I) \includegraphics[scale=0.2]{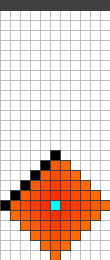}\hspace*{3cm}
II) \includegraphics[scale=0.2]{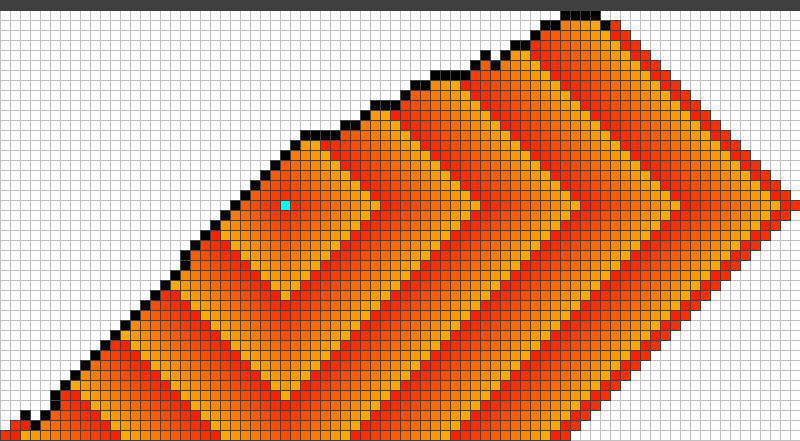}\\[5ex]
III) \includegraphics[scale=0.2]{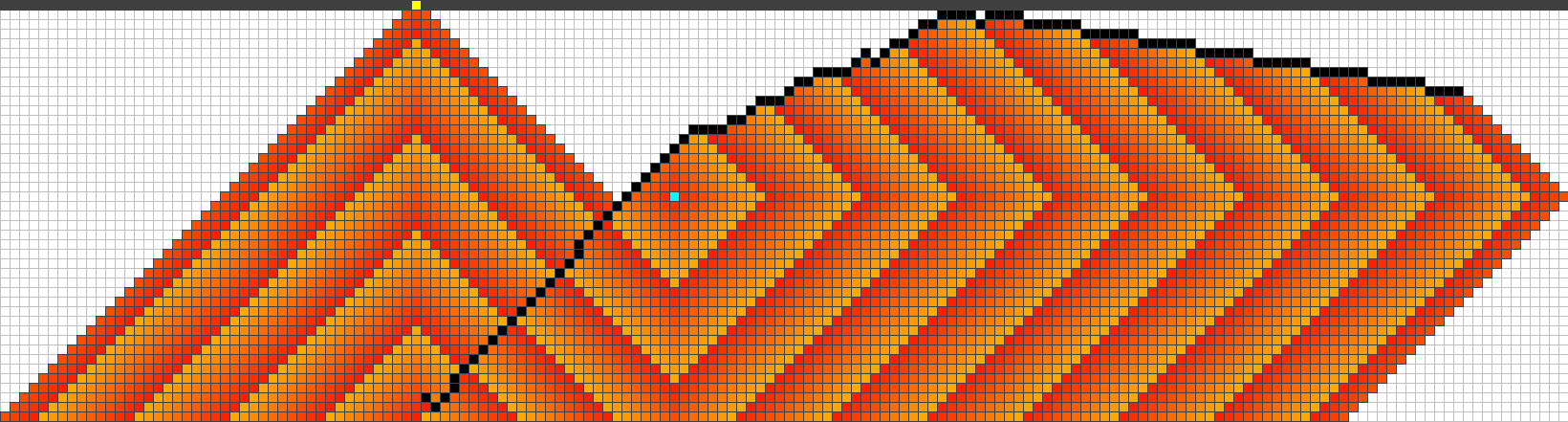}
\par\end{centering}
\caption{\label{fig:The-three-phases} Three phases of a connected strategy
protecting the highway for~$c=1.2$. I)~Reach the level of the starting 
position of the fire diagonally from the center. II)~Extend this diagonal, but 
simultaneously also try to shift the fire away from the highway on the other side
until the fire reaches the highway there. 
III)~Use the full budget to keep the fire away at this side, the fire 
\emph{runs} to the highway at the other side. 
The strategy starts 5 steps above
the origin. The fire reaches the highway after 92 steps.}
\end{figure}

If the starting position is somehow chosen \emph{too close} to the fire origin or  \emph{too close} to the highway we obtain strategies that behave in a symmetric and alternating  way as before. 
On the other hand if we somehow  start at the \emph{right} distance, the 
attained strategies suddenly performed different and a lot better.
An example of such a strategy for $c=1.2$ is given in Figure \ref{fig:The-three-phases}, the behaviour of the strategy can be subdivided into three different phases 
which will be explained below. 
In contrast to the symmetric and alternating strategy which only kept the fire for
48 steps away from the highway  with the same budget, the alternative strategy
increases this time to 92!

In general these strategies can be subdivided into three phases. 
Figure \ref{fig:The-three-phases} shows the end of each phase. 
\begin{itemize}
\item[I)] Protect a diagonal downwards until a cell at the same level as the
origin is reached. Starting $n$ cells above the origin, this requires
the protection of $n+1$ cells and this needs to be done before the
$n-$th time step, because otherwise the fire would reach that cell
first. This in turn requires $c$ to be large enough. Or the
other way round, given $1.0<c<1.5$, $n$ needs to be large enough
such that $n+1$ cells can be protected after $n$ steps $\Leftrightarrow$
$cn\geq n+1$ $\Leftrightarrow$ $n\geq\frac{1}{c-1}$.
\item[II)]  Continue the diagonal
downwards by one cell in every second step. Use the rest of the budget
to keep the fire at the other end of the barrier as far away from
the highway as possible. This procedure ends when the fire gets close to the highway. 
\item[III)] In order to protect the highway, from now on we are forced to protect
at least one cell per step at the end close to the highway. Since
protecting one cell at every step at one end and one cell at every
second step at the other end would require a budget of $c\geq1.5$,
the diagonal part of the barrier will be overrun by the fire making
it impossible to continue this end at all. So the strategy will simply
continue to hold the fire back at the upper part of the barrier until
the fire reaches the highway on the other side. Again, because the
fitness evaluation prefers fewer burning vertices close to the highway,
the slope of the part built in this phase occurs.
\end{itemize}
We will refer to this behaviour as  an \emph{ asymmetric and diagonal} strategy. Notice
that for $c=1.5$ this leads immediately to a strategy that protects
the highway infinitely. Furthermore this strategy can only be applied 
if the fire starts far enough away from the highway. The closer the budget
$c$ gets to 1.0, the more distance is required. If this distance is
not available, there seems to be no better strategy than the symmetric and alternating one. 

\subsection{Highway protection conclusion}\label{HighwayConcl-sect}

Using the evolutionary algorithm, we gained helpful insights into
the highway protection problem. Both the symmetric and alternating and the 
asymmetric and diagonal strategy are 
promising candidates for optimal solutions. 
The choice between the two alternatives seem 
to depend on the possibility of building the diagonal of phase~I). 
We found out that with one additional initial budget, the connected 
genome always run into the asymmetric and diagonal variant. 
By protecting two cells in the beginning, we can immediately 
finish the first phase by starting directly above and to the left of the
fire. Without an initial budget the strategy has to fight for reaching 
the starting level of the fire source from the left. This can only 
happen if in comparison to the budget, the source lies sufficiently 
far away from the highway.  

Considering phase III), there seems to be some room for a further 
recursive improvement.
When the fire has overcome the diagonal part of the barrier in phase II)  
it will take the direct way to the highway. For a while we shift the fire 
away from the other side by using the full budget. The barrier is build 
with a given slope; see Figure \ref{fig:The-three-phases}~III). 
But this part could have been build also with budget~$1$ along the highway.
Therefore the remaining budget can be used to protect the 
highway at the left hand side. Therefore we can consider 
the situation with a budget $c'=c-1$. We found out that in this case 
the best strategy builds a symmetric and alternating barrier directly at 
the highway.

For values $c\approx1.5$ we never observed a strategy that was able
to protect the highway infinitely long. We think that $1.5$ is the threshold.

{\bf Conjecture: } For $c<1.5$ there is no strategy that protects 
an arbitrary highway from the fire.  The best protection strategy either builds a single connected 
barrier symmetrically and alternating close to the highway  or first the
asymmetric and diagonal connected barrier strategy is applied. 
This depends on the relationship between the distance of the fire source to the highway and 
the given budget.

\section{Future work on theoretical threshold questions}\label{Future-sect} 

Besides proving and analysing the above conjecture theoretically, 
we finally would like to mention some examples for other interesting upper and lower bounds which we analogously would like to attack by goal oriented evolutionary approaches.
Similar to the subjects presented here there are other scenarios in discrete and 
continuous fire fighting settings that come along with a threshold. 
An interesting overview for such gaps is given in the CG Column by Klein and Langetepe\cite{kl-cgc-16}.
Alternatively, one might also think of the protection for different objects, also 
formalized by a set of cells. 

 Additionally, in an \emph{online} motion planning setting one can ask for 
a strategy of exploring an \emph{unknown} tree 
simultaneously by $k$-agents. 
All agents start at a the common root. 
The outgoing edges of a vertex become visible, if a vertex is visited for the first time. 
The task is to minimize the time, when all vertices of the a priori unknown tree 
have been visited by some agent. In a competitive sense one compare this 
completion time 
to the optimal completion time attained for visiting all vertices by $k$ agents for 
the fully known tree. This optimal \emph{offline} solution can be easily attained. 
There is an algorithm that is not worse than $O(k/ \log k)$ times the 
optimal completion time for any unknown tree; see \cite{fgkp-cte-04}. 
On the other hand there is an example that no online strategy can 
be better than $\Omega(\log k/ \log \log k)$ given in \cite{d-wrnm-07}. So there is a huge (exponential) gap between 
the upper and lower bound on the so-called \emph{competitive ratio}.

%


 {
\bibliographystyle{abbrv}
\bibliography{%
        EvolutionaryFireFight}
        }

\end{document}